\documentclass[10pt,twocolumn,letterpaper]{article}
\usepackage{iccv}
\usepackage{times}
\usepackage{epsfig}
\usepackage{graphicx}
\usepackage{amsmath}
\usepackage{amssymb}
\usepackage{enumitem}
\usepackage{booktabs}
\usepackage{multirow}

\usepackage[breaklinks=true,bookmarks=false]{hyperref}

\iccvfinalcopy 

\def\iccvPaperID{20} 

\ificcvfinal\fi

\begin{document}

\title{\ CLIP-Decoder : ZeroShot Multilabel Classification
using Multimodal CLIP Aligned Representations}

\author{Muhammad Ali   \hspace{0.05cm}    Salman Khan\\
Mohamed Bin Zayed University of AI\\
ABU Dhabi UAE\\
{\tt\small \{muhammad.ali, salman.khan\}@mbzuai.ac.ae}
}

\maketitle
\ificcvfinal\thispagestyle{empty}\fi

\begin{abstract}
  Multi-label classification is an essential task utilized in a wide variety of real-world applications. Multi-label zero-shot learning is a method for classifying images into multiple unseen categories for which no training data is available, while
in general zero-shot situations, the test set may include observed classes.
The CLIP-Decoder is a novel method based on the state-of-the-art ML-Decoder attention-based head. We introduce multi-modal representation learning in CLIP-Decoder, utilizing the text encoder to extract text features and the image encoder for image feature extraction.
Furthermore, we minimize semantic mismatch by aligning image and word embeddings in the same dimension and comparing their respective representations using a combined loss, which comprises classification loss and CLIP loss. This strategy outperforms other methods and we achieve cutting-edge results on zero-shot multilabel classification tasks using CLIP-Decoder. Our method achieves an absolute increase of 3.9\% in performance compared to existing methods for zero-shot learning multi-label classification tasks. Additionally, in the generalized zero-shot learning  multi-label classification task, our method shows an impressive increase of almost 2.3\%.
\end{abstract}

\section{Introduction}

Methods that performed well in multi-label classification make use of label correlation with graph neural networks \cite{multilabel_graph, multilabel_labelcorrel,chen2019multi}, and developed better loss functions, backbones, and pre-training methods \cite{Ben-Baruch2021Multi-labelLoss, multilabel_asymloss, Ridnik2021ImageNet-21KMasses, Ridnik2020TResNet:Architecture}.
The classification head and backbone are major parts of a classification network.
The spatial embedding tensor produced by the backbone is fed into the classification head which converts it to logits \cite{He2015DeepRecognition}. To perform single-label classification, global average pooling (GAP) followed by a dense or fully connected layer is commonly used \cite{Ghosh2018AdGAP:Pooling}. For multi-label classification, Zhao-Min Chen et al. \cite{ChenMulti-Label,yang2016exploit,wu2021distributionbalanced} used GAP. Attention-based heads outperform others for different scale objects \cite{Gao2020LearningRecognition}. GAP-based classifiers do not apply to zero-shot learning (ZSL), even though they are simple and give good results for classification tasks. On the other hand, attention-based classification is computationally costly and does not provide good results in ZSL settings. Recently, Tal Ridnet et al. \cite{RidnikML-Decoder:Head} came up with a work called ML-Decoder, which is based on a transformer-decoder structure with some modifications.
\begin{figure}[t].
   
    \includegraphics[scale=0.25]{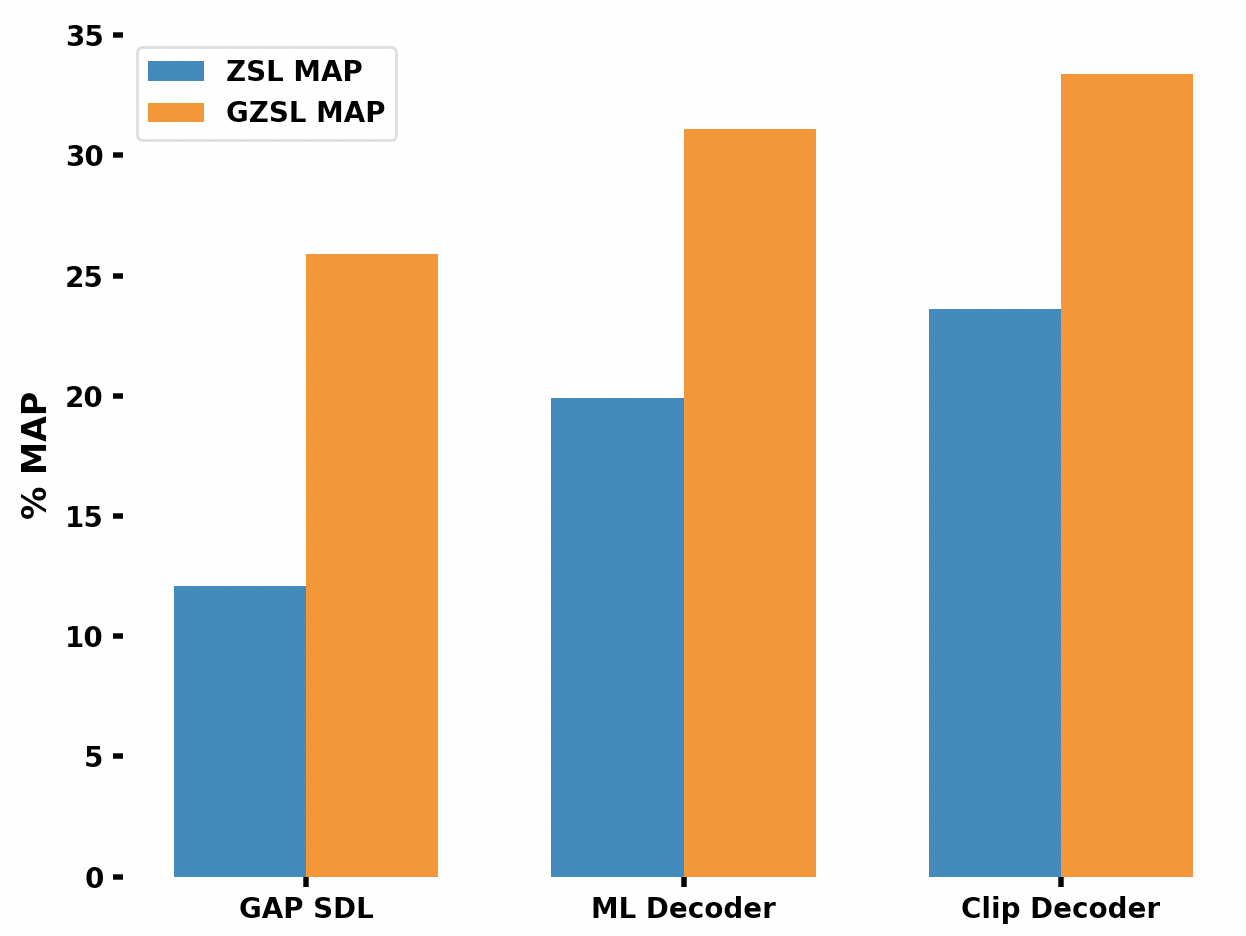} 
    \centering
    \caption{Comparison of proposed method with the state-of-the-art method in multi-label zero shot learning, ML-Decoder \cite{RidnikML-Decoder:Head}. }

    \label{fig1}
\end{figure}
We extend this work further by introducing Contrastive Language Image Pre-training(CLIP)-aligned representation learning. We generate text embedding representations and use image embeddings from TResNet to align multi-modal representations with representation learning. Our main contributions include the following:


\begin{itemize}[itemsep=-0.1em]
    \item We design prompt templates for each class in the NUS-WIDE dataset, test and verify multiple templates with their model, and select the best-performing one.
    \item We propose a multi-modal prompting approach using CLIP, aligning vision-language representations for multi-label classification in zero-shot settings.
    \item We propose the CLIP-Decoder design, leveraging a dual-modal approach to enhance transformer decoder layers by effectively fusing visual embeddings with textual information, leading to improved contextual and visual understanding. Our weighted losses establish connections between the two modalities, facilitating gradient propagation for synergy.
\end{itemize}

\section{Related Work}
A classification head and a backbone for feature extraction are two primary building blocks of classification networks \cite{abc}. Textual embeddings obtained from the text encoder and the image spatial embeddings obtained from the backbone are fed into the classification heads. We review classification heads and propose our CLIP-Decoder and its application for multi-label classification in zero-shot settings. \\
\textbf{GAP-based approach}:
    In a Global average pooling (GAP) based approach, global average pooling is used to transform spatial embeddings into a one-dimensional vector, which is fed into a fully connected layer to produce N output logits \cite{Liu2018Multi-LabelDetection,Ridnik_2021_WACV}.\\ 
\textbf{Attention-based approach}:
    Multi-label classification requires the recognition of multiple objects of varied sizes in an image. Using GAP in this context is not useful as it fails to exploit the benefits of spatial dimensions. Instead, scientists use the attention mechanism owing to its huge success in deep learning, its ability to exploit spatial information and the improvement in results \cite{liu2021query2label,Zhu_2021_ICCV}.\\ 
\textbf{ML-Decoder}:
    With the goal of reducing the computational cost, Tal Ridnik et el. suggested a new method called ML-Decoder \cite{RidnikML-Decoder:Head}. In contrast to the structure of a transformer decoder, this design is relatively simple and provides a reasonable speed-accuracy trade-off.
\section{Methodology}
Our approach for multi-label classification leverages CLIP  by designing appropriate prompts, aligning multi-modal representations, and fusing visual and textual information. The CLIP-Decoder, with transformer decoder layers, enhances contextual understanding. The multi-scale weighted joint loss optimizes classification and alignment. This strategy yields improved performance in zero-shot and generalized zero-shot learning tasks as given in Figure \ref{fig1}
\subsection{Pre-processing: Prompts design}

In the pre-processing stage, we transform labels into prompts using multiple templates.
For instance, if an image is labeled to contain the concepts, "sky", "car", and "road", the corresponding prompt would be "a photo of sky, car and road" or " a picture of sky, car, and road". 
We employed the NUS-WIDE dataset, converting input labels into multi-label prompts through preprocessing. This leveraged CLIP's internet-trained mapping, yielding significant performance advantages over word2vec, especially in zero-shot learning. For both ZSL and GZSL, we also employed different prompts  to observe their performance in various settings as given in Appendix Table 1. However, our current approach encounters difficulties when handling multiple instances of identical objects.

 


 \subsection{CLIP-Decoder: CLIP Aligned Representation learning}

\begin{figure*}[h]
  \centering
    \includegraphics[scale=0.50]{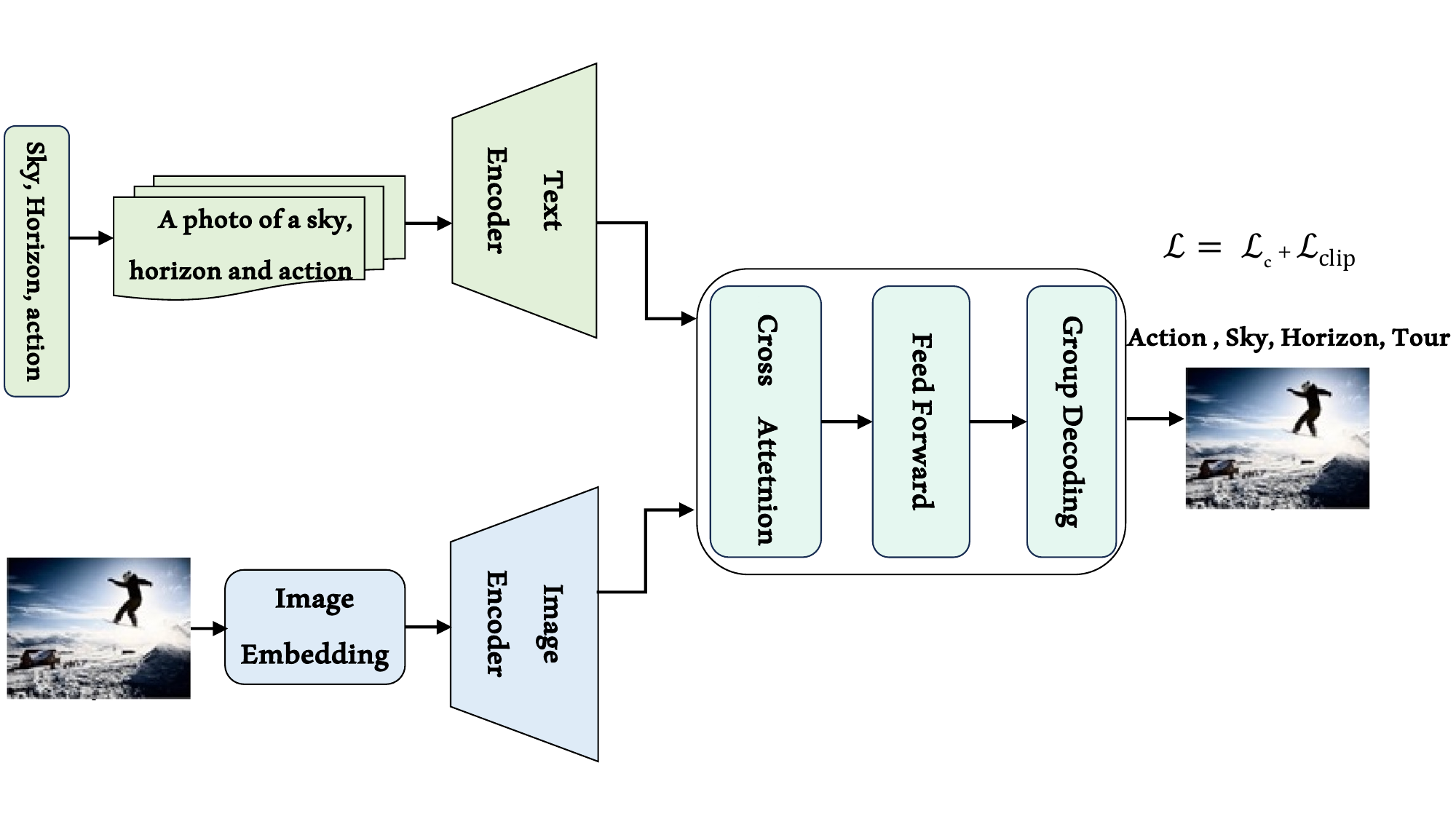} 
    \caption{Architecture of the proposed method, utilizing a TResNet image encoder for image embeddings and a CLIP text encoder for feature representations. The method integrates classification and CLIP loss for enhanced alignment}
\setlength{\abovecaptionskip}{10 pt}\label{fig_model}
\end{figure*}

To enable our model given in Figure \ref{fig_model} to learn the relationship between the image and text representations of a sample, we need to align them since they contain different information. We accomplish this by projecting both representations onto the same dimension and using the alignment loss from CLIP to align the projections. We then introduce a full-decoding version of the CLIP-Decoder for zero-shot learning (ZSL), where each label has a corresponding query.
For ZSL, we use fixed NLP-based queries, where each label is associated with a word embedding vector retrieved using a CLIP model. In the group fully-connected layer given in Figure \ref{fig_model}, we use a shared projection matrix, which transfers semantic information from the observed (training) classes to the unknown (test) classes during inference. This approach enhances the model's ability to make accurate predictions for novel classes in ZSL and improves the overall performance of the model. Inspired by \cite{RidnikML-Decoder:Head} with full-decoding (g = 1), each query checks the existence of a single class. With group decoding, each query checks the existence of several classes

\begin{equation} \setcounter{equation}{1}
    Attention(Q, K, V) = softmax(\frac{QK^T}{\sqrt{d_k}})V \label{eq:eq1}
\end{equation}
The query vector is denoted by $Q$, the key vector is denoted by $K$, and the value vector is denoted by $V$. Because the output is scaled by the dimension $d$, it is given in  the term scaled dot product.
 CLIP-Decoder attention mechanism relies on the similarity of dot products between vectors \eqref{eq:eq1}, Where $Q$, $K$ are input queries. Because NLP word embeddings maintain this dot-product semantic similarity \cite{smith2019contextual}, it is more likely that the hidden labels will correspond to the decoder's most similar keys and values. CLIP-Decoder with a shared projection matrix also supports a variable number of input queries.
In contrast to Generalized ZSL (GZSL), which does inference on the union of the visible and unseen sets of labels, ZSL trains entirely on seen labels and performs inference on the unseen classes.
To summarise, the CLIP-Decoder receives image embedding as input from the TResNet image encoder. We translate class names into prompts and then pass them through the CLIP text encoder to produce representations for each class, image embeddings are encoded using a TResNet image encoder. 
We obtain the CLIP-Loss by matching the input CLIP-aligned embeddings with image embeddings from the Image-encoder block.
Further detail is present in the Appendix under Alignment loss.
\subsubsection{Alignment Loss}

Mathematically we can write $L$ as 
 \begin{equation}
    L =\alpha  L_{clip} + \beta L_{c}\label{eq:eq2}
  \end{equation} 
\eqref{eq:eq2} shows the loss formulation where $L_{clip}$ is the CLIP alignment loss while $L_{c}$ indicates the cross entropy classification loss. 
The CLIP-based alignment loss $L_{clip}$ leverages the CLIP model's ability to embed both the visual embeddings and the text embeddings into a shared embedding space by computing cosine similarity.

\section{Experiments}
 {%

\begin{table*}[!htbp]
\centering

\label{tab:my-table}
\begin{tabular}{@{}ccccccc@{}}
\toprule
\textbf{Dataset (NUS-WIDE)} & \textbf{$L_{c}$} & \textbf{$L_{clip}$} & \textbf{Prompts} & \textbf{mAP} & \textbf{F1 k=3} & \textbf{F1 k=5} \\ \midrule
\multirow{2}{*}{CLIP-Decoder ZSL} & \checkmark &\checkmark & \checkmark & \textbf{33.43} & \textbf{34.80} & \textbf{31.10} \\
 & \checkmark & & \checkmark & 33.25 & 32.07 & 28.80 \\ \midrule
\multirow{2}{*}{CLIP-Decoder GZSL} & \checkmark & \checkmark & \checkmark & \textbf{23.80} & \textbf{24.87} & \textbf{27.54} \\
 & \checkmark & & \checkmark & 23.60 & 23.40 & 25.83 \\ \bottomrule
\end{tabular}
\caption{Multi-label ZSL and GZSL in terms of mAP as well as F1 score (\textbf{k} $\in \{3, 5\}$).}
\end{table*}

In order to evaluate our method, we used the NUS-WIDE data-set \cite{nus-wide-civr09}  which is a widely used  benchmark for  multi-label ZSL tasks.  We evaluated our model for multi-label zero-shot(MZSL) as well as generalized zero-shot(GZSL) classification tasks using mean Average Precision(mAP) and F1 score at top-K predictions, where $k$ is 3 and 5.
TResNetM is used for image feature extraction from multi-label images with a 224 image resolution. CLIP-aligned embeddings are used for text features, with a 512-d embedding size, 56-batch size, and 10-3 learning rate\cite{RadfordLearningSupervision}. 

\subsection{Training and Testing}

 During training, the network was trained on seen classes to learn their semantic space and map relevant semantic information to their corresponding classes. During evaluation, the network used non-trainable prompts to predict unseen classes, achieving improved results compared to using trainable prompts.

We conducted experiments in two settings: zero-shot learning (ZSL) and generalized zero-shot learning (GZSL). We initially used fixed embeddings (word2vec and CLIP-based prompts) during training and then employed representation alignment with a joint training loss to further enhance performance.
The joint loss combined a clip loss and a classification loss, with weights tuned through cross-validation.

In the ZSL setting, CLIP-Decoder with CLIP-embeddings outperformed using word2vec, and the joint loss approach led to much-improved results. The experiments were repeated for GZSL, where seen and unseen classes were combined, and the CLIP-aligned joint loss provided the best results. By removing the bias towards seen classes, CLIP-Decoder achieved state-of-the-art performance for both ZSL and GZSL, demonstrating improved classification for both seen and unseen classes. For training, we use 920 classes with labels while for testing we use 81 unseen labels to predict the unknown labels. For inference we input images and generate corresponding labels for each images. 

\subsection{Ablation Study Results}
In this section, we evaluate our method for both ZSL and GZSL settings. We use image embeddings from the TResNet image encoder as it is more reliable, efficient, and effective.
We first provide word2vec fixed embeddings in CLIP-Decoder which is given in Table \ref{tab:tab1} as without any tick mark in CLIP-embedding. Then we replace  CLIP embeddings inplace of word2vec which gave us an increase in mAP values as well as F1 Score values for k=3 and k=5 as given in Table \ref{tab:tab1}, we only use $\mathrm{L_c}$ here. In ZSL evaluations with 81 unseen and 925 seen classes, the shift to joint loss ($\mathrm{L_c}$ and $\mathrm{L_c}$) training yields improved results.
For both ZSL and GZSL, we also design different prompts to observe their performance in various settings and select the one with better results.
Keeping previous settings of ZSL, we evaluate our approach in GZSL, where we add the seen and unseen classes together and try to predict the unseen classes out of them. As we can see, we got an overall increase in MAP of 3.7\% with good competitive F1 score values for k = 3 and k = 5, as given in  Table \ref{tab:tab2}.
In GZSL, we train just like in ZSL. But, in the testing or evaluation phase, we test from a whole set of seen and unseen classes together to assess our model's ability to relate relevant images present in the combined set of seen and unseen classes with corresponding labels.
\begin{table*}[!htbp]
\centering
\resizebox{0.67\textwidth}{!}{%
\begin{tabular}{@{}cccccc@{}}
\toprule
\textbf{Method}                        & \textbf{Task} & \textbf{mAP} & \textbf{F1 k=3} & \textbf{F1 k=5} \\ 
\midrule
\multirow{2}{*}{Attention per Cluster \cite{vyas2020fast}} & GZSL          & 2.6          & 6.4              & 7.7              \\
                                       & ZSL           & 12.9         & 24.6             & 22.9             \\ 
\midrule
\multirow{2}{*}{LESA \cite{chen2019multi}}                  & GZSL          & 5.6          & 14.4             & 16.8             \\
                                       & ZSL           & 19.4         & 31.6             & 28.7             \\
\midrule
\multirow{2}{*}{BiAM \cite{yang2023speech}}                  & GZSL          & 9.3          & 16.1             & 19               \\
                                       & ZSL           & 26.3         & 33.1             & 30.7             \\
\midrule
\multirow{2}{*}{SDL \cite{bencohen2021semantic}}                   & GZSL          & 12.1         & 18.5             & 27.8             \\
                                       & ZSL           & 25.9         & 30.5             & 21               \\
\midrule
\multirow{2}{*}{ML Decoder \cite{ridnik2023ml}}            & GZSL          & 19.9         & 23.3             & 26.1             \\
                                       & ZSL           & 31.1         & 34.1             & 30.8             \\ 
\midrule
\multirow{2}{*}{CLIP-Decoder}          & \textbf{GZSL}          & \textbf{23.8}       & \textbf{24.87}             & \textbf{27.54}            \\
                                       & \textbf{ZSL}           & \textbf{33.4}        & \textbf{34.8}             & \textbf{31.1}            \\ 
\bottomrule
\end{tabular}}
\caption{State-of-the-art comparison for ZSL and GZSL on the NUS-WIDE dataset.}
\label{tab:tab2}
\end{table*}

\subsection{State-of-the-art Comparison}
Table \ref{tab:tab2} compares the State-of-the-art methods \cite{RidnikML-Decoder:Head} for ZSL and GZSL. Table 3 compares state-of-the-art methods for zero-shot and generalized zero-shot (ZSL) label prediction. CLIP-Decoder outperforms SDL in GZSL, while BiAm performs better in ZSL, indicating biased models. Our work, shows improved results in both ZSL and GZSL. \\
\textbf{ Results and Evaluation}:
We compare our technique with existing methods in Table \ref{tab:tab2}, using mAP values and F1 score as performance metrics. Our approach shows a 2.33\% and 3.7\% absolute gain in mAP for ZSL and GZSL, respectively, while maintaining competitive F1 score values. Previous methods prioritize good results at the expense of GZSL performance, except for ML-Decoder. As given in Table \ref{tab:tab2} Semantic Diversity learning(SDL) shows the better performance in generalized zeroshot learning (GZSL)in comparison with Descriminative Region based Multilabel Zeroshot learning(BiAM), while performance in zero shot settings(ZSL) degrades. The ML Decoder method improves existing results in all metrics, while our CLIP module further enhances mAP for both ZSL and GZSL cases. The introduction of CLIP representation learning on top of ML-Decoder give us consistent increases in mAP values, as shown in Table \ref{tab:tab2}. CLIP alignment is useful because the CLIP model is trained on large amounts of image and text pairs, allowing it to predict relevant image labels given text input. 
Figure \ref{fig3} shows the multilabel classification performed using CLIP-Decoder.

\begin{figure}[h]
   \centering
    \includegraphics[scale=0.40]{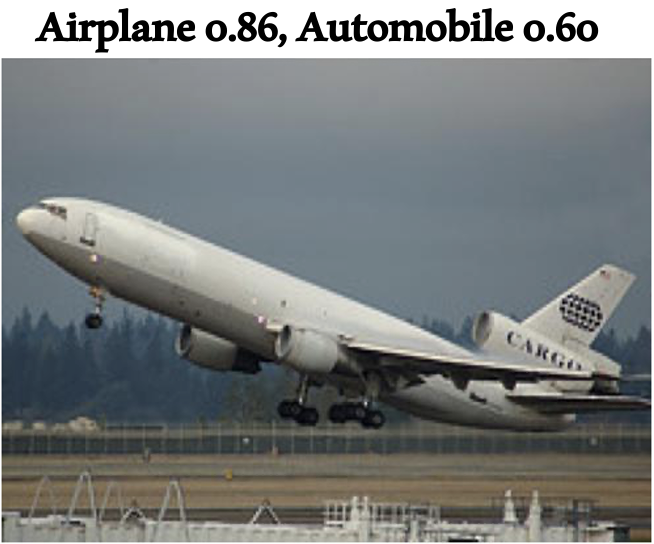} 
    \caption{ Multi-label classification in GZSL.}
    \label{fig3}
\end{figure}

\section{Conclusion}
We introduce the CLIP-Decoder, which improves multi-label classification in a zero-shot context by aligning image and text representations using representation learning. We project text and image representations onto the same dimension and use CLIP's alignment loss to align them. This enhances multi-modal representation learning, resulting in better synergy between vision and language modalities. Our approach outperforms existing state-of-the-art methodologies in both zero-shot and generalized zero-shot contexts. We intend to extend this approach to other domains such as zero-shot action recognition and detecting new cancer types in multiomics frameworks. 

{\small
\bibliographystyle{ieee_fullname}
\bibliography{egbib}

\begin{thebibliography}{10}\itemsep=-1pt

\bibitem{multilabel_asymloss}
Emanuel~Ben Baruch, Tal Ridnik, Nadav Zamir, Asaf Noy, Itamar Friedman, Matan Protter, and Lihi Zelnik{-}Manor.
\newblock Asymmetric loss for multi-label classification.
\newblock {\em CoRR}, abs/2009.14119, 2020.

\bibitem{Ben-Baruch2021Multi-labelLoss}
Emanuel Ben-Baruch, Tal Ridnik, Itamar Friedman, Avi Ben-Cohen, Nadav Zamir, Asaf Noy, and Lihi Zelnik-Manor.
\newblock Multi-label classification with partial annotations using class-aware selective loss.
\newblock In {\em Proceedings of the IEEE/CVF Conference on Computer Vision and Pattern Recognition}, pages 4764--4772, 2022.

\bibitem{bencohen2021semantic}
Avi Ben-Cohen, Nadav Zamir, Emanuel~Ben Baruch, Itamar Friedman, and Lihi Zelnik-Manor.
\newblock Semantic diversity learning for zero-shot multi-label classification, 2021.

\bibitem{multilabel_graph}
Zhao{-}Min Chen, Xiu{-}Shen Wei, Peng Wang, and Yanwen Guo.
\newblock Multi-label image recognition with graph convolutional networks.
\newblock {\em CoRR}, abs/1904.03582, 2019.

\bibitem{ChenMulti-Label}
Zhao{-}Min Chen, Xiu{-}Shen Wei, Peng Wang, and Yanwen Guo.
\newblock Multi-label image recognition with graph convolutional networks.
\newblock {\em CoRR}, abs/1904.03582, 2019.

\bibitem{multilabel_labelcorrel}
Zhao-Min Chen, Xiu-Shen Wei, Xin Jin, and Yanwen Guo.
\newblock Multi-label image recognition with joint class-aware map disentangling and label correlation embedding.
\newblock {\em 2019 IEEE International Conference on Multimedia and Expo (ICME)}, pages 622--627, 2019.

\bibitem{chen2019multi}
Zhao-Min Chen, Xiu-Shen Wei, Xin Jin, and Yanwen Guo.
\newblock Multi-label image recognition with joint class-aware map disentangling and label correlation embedding.
\newblock In {\em 2019 IEEE International Conference on Multimedia and Expo (ICME)}, pages 622--627. IEEE, 2019.

\bibitem{nus-wide-civr09}
Tat-Seng Chua, Jinhui Tang, Richang Hong, Haojie Li, Zhiping Luo, and Yan-Tao Zheng.
\newblock Nus-wide: A real-world web image database from national university of singapore.
\newblock In {\em Proc. of ACM Conf. on Image and Video Retrieval (CIVR'09)}, Santorini, Greece., July 8-10, 2009.

\bibitem{Gao2020LearningRecognition}
Bin-Bin Gao and Hong-Yu Zhou.
\newblock {Learning to Discover Multi-Class Attentional Regions for Multi-Label Image Recognition}.
\newblock {\em IEEE Transactions on Image Processing}, 30:5920--5932, 7 2020.

\bibitem{Ghosh2018AdGAP:Pooling}
Arna Ghosh, Biswarup Bhattacharya, and Somnath Basu~Roy Chowdhury.
\newblock {AdGAP: Advanced Global Average Pooling}.
\newblock {\em Proceedings of the AAAI Conference on Artificial Intelligence}, 32(1):8081--8082, 4 2018.

\bibitem{He2015DeepRecognition}
Kaiming He, Xiangyu Zhang, Shaoqing Ren, and Jian Sun.
\newblock {Deep Residual Learning for Image Recognition}.
\newblock {\em Proceedings of the IEEE Computer Society Conference on Computer Vision and Pattern Recognition}, 2016-December:770--778, 12 2015.

\bibitem{abc}
Kaiming He, Xiangyu Zhang, Shaoqing Ren, and Jian Sun.
\newblock Deep residual learning for image recognition.
\newblock In {\em 2016 IEEE Conference on Computer Vision and Pattern Recognition (CVPR)}, pages 770--778, 2016.

\bibitem{liu2021query2label}
Shilong Liu, Lei Zhang, Xiao Yang, Hang Su, and Jun Zhu.
\newblock Query2label: A simple transformer way to multi-label classification, 2021.

\bibitem{Liu2018Multi-LabelDetection}
Yongcheng Liu, Lu Sheng, Jing Shao, Junjie Yan, Shiming Xiang, and Chunhong Pan.
\newblock {Multi-Label Image Classification via Knowledge Distillation from Weakly-Supervised Detection}.
\newblock {\em MM 2018 - Proceedings of the 2018 ACM Multimedia Conference}, pages 700--708, 9 2018.

\bibitem{RadfordLearningSupervision}
Alec Radford, Jong~Wook Kim, Chris Hallacy, Aditya Ramesh, Gabriel Goh, Sandhini Agarwal, Girish Sastry, Amanda Askell, Pamela Mishkin, Jack Clark, et~al.
\newblock Learning transferable visual models from natural language supervision.
\newblock In {\em International Conference on Machine Learning}, pages 8748--8763. PMLR, 2021.

\bibitem{Ridnik2021ImageNet-21KMasses}
Tal Ridnik, Emanuel Ben-Baruch, Asaf Noy, and Lihi Zelnik.
\newblock Imagenet-21k pretraining for the masses.
\newblock In J. Vanschoren and S. Yeung, editors, {\em Proceedings of the Neural Information Processing Systems Track on Datasets and Benchmarks}, volume~1, 2021.

\bibitem{Ridnik2020TResNet:Architecture}
Tal Ridnik, Hussam Lawen, Asaf Noy, Emanuel~Ben Baruch, Gilad Sharir, and Itamar Friedman.
\newblock {TResNet: High Performance GPU-Dedicated Architecture}.
\newblock {\em Proceedings of the IEEE/CVF Winter Conference on Applications of Computer Vision.}, 3 2021.

\bibitem{Ridnik_2021_WACV}
Tal Ridnik, Hussam Lawen, Asaf Noy, Emanuel Ben~Baruch, Gilad Sharir, and Itamar Friedman.
\newblock Tresnet: High performance gpu-dedicated architecture.
\newblock In {\em Proceedings of the IEEE/CVF Winter Conference on Applications of Computer Vision (WACV)}, pages 1400--1409, January 2021.

\bibitem{RidnikML-Decoder:Head}
Tal Ridnik, Gilad Sharir, Avi Ben{-}Cohen, Emanuel~Ben Baruch, and Asaf Noy.
\newblock Ml-decoder: Scalable and versatile classification head.
\newblock {\em CoRR}, abs/2111.12933, 2021.

\bibitem{ridnik2023ml}
Tal Ridnik, Gilad Sharir, Avi Ben-Cohen, Emanuel Ben-Baruch, and Asaf Noy.
\newblock Ml-decoder: Scalable and versatile classification head.
\newblock In {\em Proceedings of the IEEE/CVF Winter Conference on Applications of Computer Vision}, pages 32--41, 2023.

\bibitem{smith2019contextual}
Noah~A Smith.
\newblock Contextual word representations: A contextual introduction.
\newblock {\em arXiv preprint arXiv:1902.06006}, 2019.

\bibitem{vyas2020fast}
Apoorv Vyas, Angelos Katharopoulos, and Fran{\c{c}}ois Fleuret.
\newblock Fast transformers with clustered attention.
\newblock {\em Advances in Neural Information Processing Systems}, 33:21665--21674, 2020.

\bibitem{wu2021distributionbalanced}
Tong Wu, Qingqiu Huang, Ziwei Liu, Yu Wang, and Dahua Lin.
\newblock Distribution-balanced loss for multi-label classification in long-tailed datasets, 2021.

\bibitem{yang2016exploit}
Hao Yang, Joey~Tianyi Zhou, Yu Zhang, Bin-Bin Gao, Jianxin Wu, and Jianfei Cai.
\newblock Exploit bounding box annotations for multi-label object recognition, 2016.

\bibitem{yang2023speech}
Yuhang Yang, Haihua Xu, Hao Huang, Eng~Siong Chng, and Sheng Li.
\newblock Speech-text based multi-modal training with bidirectional attention for improved speech recognition.
\newblock In {\em ICASSP 2023-2023 IEEE International Conference on Acoustics, Speech and Signal Processing (ICASSP)}, pages 1--5. IEEE, 2023.

\bibitem{Zhu_2021_ICCV}
Ke Zhu and Jianxin Wu.
\newblock Residual attention: A simple but effective method for multi-label recognition.
\newblock In {\em Proceedings of the IEEE/CVF International Conference on Computer Vision (ICCV)}, pages 184--193, October 2021.

\end{thebibliography}
}
\end{document}